\documentclass[1p]{elsarticle}

\usepackage[utf8]{inputenc}

\usepackage{amsmath,amssymb,graphicx,color,subfig,url,rotating,xspace,synttree,xcolor,figparameters}
\usepackage{textcomp} 
\usepackage[normalem]{ulem}

\newcommand{\taaable}[0]{{\sc Taaable}\xspace}

\def\indu{\ifmmode \mathcal{INDU} \else $\indu$ \fi}

\def\ing{\mathcal{I}}

\def\brop{\ifmmode\dotplus\else$\brop$\fi}

\def\regor{\textbar\xspace}

\def\aa{A}
\def\bb{B}

\def\mlm#1{\text{\begin{tabular}{c}#1\end{tabular}}}

\def\mlm#1{\text{\begin{tabular}{c}#1\end{tabular}}} 

\def\ensuremathxspace#1{\ifmmode#1\else$#1$\xspace\fi}

\def\relQCN#1{\relQCNsa{\{#1\}}}
\def\relQCNsa#1{\ensuremath{\mathrel{\text{#1}}}} 
\def\alrel#1{\relQCN{#1}} 
\def\alrelnb#1{\relQCNsa{#1}}

\def\indu{\ensuremathxspace{\mathcal{INDU}}}
\def\relQCNindu#1#2{\relQCN{#1}\ensuremath{^{\text{#2}}}}
\def\relQCNsaindu#1#2{\relQCNsa{#1\ensuremath{^{\text{#2}}}}}

\journal{Information Systems}

\begin{document}

\begin{frontmatter}

   \title{Automatic Case Acquisition From Texts\\for Process-Oriented Case-Based Reasoning}
   
\author[ul,cnrs,inria]{Valmi Dufour-Lussier\corref{cor1}}
\ead{valmi.dufour@loria.fr}
\cortext[cor1]{Corresponding author.}

\author[us,cnrs,inria]{Florence Le Ber}
\ead{florence.leber@engees.unistra.fr}

\author[ul,cnrs,inria]{Jean Lieber}
\ead{jean.lieber@loria.fr}

\author[ul,cnrs,inria]{Emmanuel Nauer}
\ead{emmanuel.nauer@loria.fr}

\address[ul]{Université de Lorraine, LORIA, UMR 7503 --- Vand\oe{}uvre-lès-Nancy, F-54506, France}
\address[cnrs]{CNRS, LORIA, UMR 7503 --- Vand\oe{}uvre-lès-Nancy, F-54506, France}
\address[inria]{Inria --- Villers-lès-Nancy, F-54602, France}
\address[us]{ENGEES, LHYGES, UMR 7517 --- Strasbourg, F-67000, France}
   
   \begin{abstract}
   	This paper introduces a method for the automatic acquisition of a rich case representation from free text for process-oriented case-based reasoning.
		Case engineering is among the most complicated and costly tasks in implementing a case-based reasoning system.
		This is especially so for process-oriented case-based reasoning, where more expressive case representations are generally used and,
		in our opinion, actually required for satisfactory case adaptation.
		In this context, the ability to acquire cases automatically from procedural texts is a major step forward in order to reason on processes.
		We therefore detail a methodology that makes case acquisition from processes described as free text possible,
		with special attention given to assembly instruction texts.
		This methodology extends the techniques we used to extract actions from cooking recipes.
		We argue that techniques taken from natural language processing are required for this task,
		and that they give satisfactory results.
		An evaluation based on our implemented prototype extracting workflows from recipe texts is provided.
   \end{abstract}
\end{frontmatter}

\section{Introduction}
\label{sec:intro}
This paper introduces a method for the automatic acquisition of a rich case representation from free text for process-oriented case-based reasoning.
Case-based reasoning (CBR)~\cite{lopez06ker} is a reasoning paradigm used to solve problems by retrieving similar problem--solution pairs from a case base and adapting the solutions. Because case engineering is a complicated and costly process, automatic case acquisition from text has been a major focus of research in the last 15 years. Methods used in textual CBR vary widely in their scope and their use of natural language processing (NLP) methods, depending on the intended applications.
At one end of the spectrum, a text is seen as a bag-of-words and is mostly represented as a term vector. Weights are assigned according to the relative frequency of the terms in a text versus its frequency in all the texts of the case base, possibly taking synonymy into account. This method does not allow for sophisticated inferences, such as adaptation of highly structured cases, taking into account domain knowledge.
At the other end, Gupta and Aha~\cite{gupta04flairs} propose using a full-fledged NLP solution to translate free text into predicate logic. While such a system would indeed allow for very sophisticated adaptation (including, if combined with a natural language generation system, unlimited possibilities in terms of textual adaptation), existing NLP systems translating from text to logic are far from being accurate.

In this paper, we argue in favour of using limited NLP techniques to extract, from procedural texts, the verbs, their complements and their relevant modifiers, in order to acquire rich case representations for processes. We are more open to linguistic-inspired techniques than approaches to textual CBR based on information retrieval and even information extraction. By limiting ourselves to procedural texts (an important but not very varied type of text) and trying to extract only domain-relevant meaning from it, we are able to achieve encouraging results.
The approach presented herein is a generalisation of our previous work in defining a case representation for recipes and implementing software to acquire cases automatically from a recipe book~\cite{dufour10iccbr}. This domain shows numerous examples where case representations richer than a vector space and NLP are a necessity.


Some background knowledge is required to understand certain choices made in designing this extraction process. This is introduced briefly in sections~\ref{sec:form} and~\ref{sec:appl}. Section~\ref{sec:form} presents two formalisms that are used in process case representations: workflows and qualitative algebras. Some shortcomings of both are pointed out, which justifies making sure that the proposed method can extract both workflows and qualitative constraints from text. Additionally, two running examples are introduced: a cooking recipe and a scientific experimental protocol. Section~\ref{sec:appl} presents different usages that are expected for the case representation, and shows which effect they have on the extraction process. More specifically, two approaches for textual case adaptation are presented: one based on ``grafting'', and one based on belief revision theory.

Section~\ref{sec:extr} is the core of the paper. It describes in details the process through which case representations are extracted from texts. A text is analysed in many passes. All but the last we use implement fairly typical NLP methods, and are detailed in subsection~\ref{sec:stand}. The last one, introduced in subsection~\ref{sec:anaph}, is an anaphoric reference solver that we designed specifically for procedural texts. Subsection~\ref{sec:acquis} explains how a structured case representation is engineered from the information extracted in the previous steps. This is sufficient to allow for adaptation of the formalised cases, but additional concerns that must be kept it mind to facilitate the reuse of the actual text are introduced in subsection~\ref{sec:adapt}. All the examples in this section come from recipes, but we show in subsection~\ref{sec:chem} that the method presented is applicable to different types of texts by applying them to the scientific protocol.

Section~\ref{sec:eval} presents a first evaluation of the extraction process by comparing the case representations obtained automatically against a gold standard. Section~\ref{sec:rw} and~\ref{sec:concl} finally discuss related and future work.

\section{Formalisms for process representation}
\label{sec:form}
The most important method used to model cases representing processes is workflows. For instance, all the papers presented at the first Process-oriented Case-based Reasoning workshop~\cite{pocbr11} were based on workflows, except two, that used workflow-like graphs.
Workflows efficiently model the actions required to complete a process: a workflow defines a partial order over the actions and offers the possibility to express options (disjunctions), simultaneity (conjunctions) and repetition of actions (loops).
There exist different formal languages in which workflow knowledge can be represented.
In this paper, UML Activity Diagrams are used.

\begin{figure}
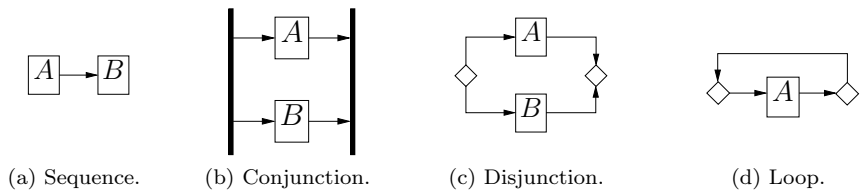

  {
  \def\leve#1#2{\raisebox{#1mm}{#2}}
  \def\mef#1{{\huge$#1$}} 
  \def\aa{\mef{A}}
  \def\bb{\mef{B}}
  \begin{center}
  \subfloat[Sequence.]{\label{fig:wf-seq}
      ~~\leve{8}{\resizebox{0.10\textwidth}{!}{\input{sequence.t.tex}}}~~}
  \qquad
  \subfloat[Conjunction.]{\label{fig:wf-conj}
      ~~\resizebox{0.125\textwidth}{!}{\input{conjunction.t.tex}}~~}
  \qquad
  \subfloat[Disjunction.]{\label{fig:wf-disj}
      ~~\leve{3}{\resizebox{0.15\textwidth}{!}{\input{disjunction.t.tex}}}~~}
  \qquad
  \subfloat[Loop.]{\label{fig:wf-loop}
      ~~\leve{5}{\resizebox{0.15\textwidth}{!}{\input{loop.t.tex}}}~~}
  \end{center}
  }
	\caption{Workflow patterns.}
	\label{fig:wf-cfs}
\end{figure}

The most basic control flow structure is the sequence (shown in figure~\ref{fig:wf-seq}), which means that a task $\bb$ is ready to begin execution as soon as a task $\aa$ is finished. Other control statements are the fork, indicating the concurrent execution of workflows, the join, synchronising them, the decision, leading to the exclusive execution of one workflow from a set, and the merge, ending such a conditional execution~\cite{aalst03dpd}. The fork and join control statements are used to create a conjunction control structure (figure~\ref{fig:wf-conj}), while the decision and merge statements can be used to create either a disjunction (figure~\ref{fig:wf-disj}) or a loop structure (figure~\ref{fig:wf-loop}).

Workflows have some limitations, apparent in the representation of the biochemistry procedure shown in figure~\ref{fig:txt-biochem}. Temporal aspects that are important in certain types of processes cannot be represented, including duration of actions (e.g. the fact that the powder is refluxed for 3 hours in step~7d of the procedure) and fine temporal relations between actions (e.g. the constraint that the powder must be dried \emph{immediately} after centrifugation in step~7e).

\begin{figure}
	\begin{center}
   	\input{txt-biochem}
   	\caption{Excerpts from a scientific experimental protocol for preparing zymosan~\cite{pillemer56jem}.}
   	\label{fig:txt-biochem}
	\end{center}
\end{figure}

\begin{figure}
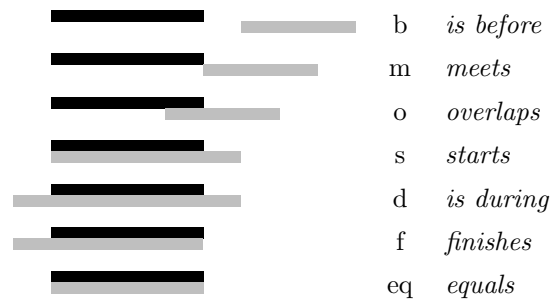

  {
  \def\intx{$\rule[1.5mm]{20mm}{1.5mm}$}
  \def\inty#1#2{{\textcolor{gray!50}{\hspace{#1mm}$\rule[0mm]{#2mm}{1.5mm}$}}}
  \def\ligne#1#2#3#4{\makebox[0mm][l]{\intx}\inty{#1}{#2} & $\alrelnb{#3}$ & \emph{#4}}
  \def\rien#1{#1} 
  %
  \begin{center}
    \begin{tabular}{l c l}
      \ligne{25}{15}{b}{is \rien{b}efore}
      \\[1.5mm]
      \ligne{20}{15}{m}{\rien{m}eets}
      \\[1.5mm]
      \ligne{15}{15}{o}{\rien{o}verlaps}
      \\[1.5mm]
      \ligne{0}{25}{s}{\rien{s}tarts}
      \\[1.5mm]
      \ligne{-5}{30}{d}{is \rien{d}uring}
      \\[1.5mm]
      \ligne{-5}{25}{f}{\rien{f}inishes}
      \\[1.5mm]
      \ligne{0}{20}{eq}{\rien{eq}uals}
    \end{tabular}
  \end{center}
}

  \caption{Allen interval algebra's relations.
  \label{fig:allen}}
\end{figure}

Qualitative algebras over intervals can be used, with actions reified as intervals, to express fine temporal constraints between actions.
Allen algebra~\cite{allen81ijcai,allen83cacm}, for instance,
 provides 13 relations corresponding to all the possible order relations between the beginnings and the ends of two intervals.
7 relations are illustrated in figure~\ref{fig:allen},
 the 6 others being the inverse of the first 6 (\alrelnb{eq} is symmetric).

\indu~\cite{pks99atai} extends Allen algebra with relations over interval durations:
 each Allen relation is augmented with either \alrelnb{$^<$}, \alrelnb{$^=$} or~\alrelnb{$^>$}.
Since not all combinations are possible (e.g. \alrelnb{eq} can only be combined with \alrelnb{$^=$}),
 this actually yields 25 base relations.
\alrelnb{r$^?$} is a shortcut notation for Allen relation \alrelnb{r} with any duration relation,
 and \alrelnb{$?^\text{s}$} is a shortcut for any Allen relation with duration relation \alrelnb{s}
  (e.g.,
  $\relQCNindu{m}{$?$}
   =
   \{\relQCNsaindu{m}{$<$},\relQCNsaindu{m}{$=$},\relQCNsaindu{m}{$>$}\}$;
  $\relQCNindu{d}{$?$}
   =
   \{\relQCNsaindu{d}{$<$}\}$;
  $\relQCNsaindu{?}{$=$}
   =
   \{\relQCNsaindu{b}{$=$},\relQCNsaindu{m}{$=$},\relQCNsaindu{o}{$=$},\relQCNsaindu{eq}{$=$},\relQCNsaindu{oi}{$=$},\relQCNsaindu{mi}{$=$},\relQCNsaindu{bi}{$=$}\}$.
Using intervals that correspond to fixed durations rather than to actions with \indu makes it possible, up to a certain point, to express durations:
 for instance, ``$\text{cook}\alrelnb{$?^=$}\text{15 min}$'' would represent cooking for 15 minutes.
On the other hand, no qualitative algebra offers a satisfactory way to express either disjunctions or loops,
 making them insufficiently expressive to represent certain abstract classes of processes.

Qualitative knowledge bases are often represented as graphs: the vertices are the intervals, and the arcs are the relations. 
Relations are shown as sets, which represent a disjunction of base relations. For instance, $i \xrightarrow{\alrel{b,m}^<} j$ indicates that interval $i$ is shorter than interval $j$ ($^<$ relation), and either $i$ occurs entirely before $j$ (\alrelnb{b} relation), or the end of $i$ coincides with the beginning of $j$ (\alrelnb{m} relation).

In the rest of this paper, we will present the case representation extraction from text with the perspective that it should primarily generate workflows. Because workflows are nor always sufficiently expressive though, the formalism will be switched to \indu whenever information that cannot be expressed with workflows is encountered. Ultimately, we consider that process-oriented CBR systems should be able to deal with both workflows and qualitative algebras.

We mostly illustrate our technique using cooking recipes because they are readily understandable by almost anyone and their domain terminology needs no definition. 

\begin{figure}
	\begin{center}
   	\input{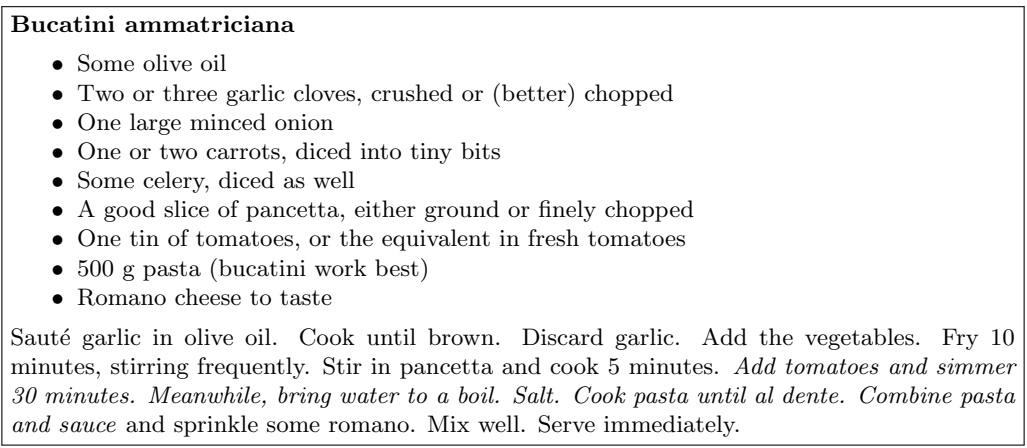}
   	\caption{A recipe for cooking bucatini ammatriciana---the italicised part is represented as a workflow in figure~\ref{fig:form-eg-wf} and as a set of qualitative constraints in figure~\ref{fig:form-eg-qa}}
   	\label{fig:txt-recipe}
	\end{center}
\end{figure}

\begin{figure}
	\begin{center}
  {
  \def\mef#1{{\huge$#1$}} 
  \def\meft#1{\text{#1}} 
  \def\aa{\meft{add}}
  \def\bb{\meft{simmer}}
  \def\cc{\meft{\mlm{bring \\ water to \\ a boil}}}
  \def\dd{\meft{\mlm{cook \\ bucatini}}}
  \def\ee{\meft{\emph{not al dente}}}
  \def\ff{\meft{combine}}
  \def\tomatoes{\meft{tomatoes}}
  \def\bucatini{\meft{bucatini}}
  \resizebox{\textwidth}{!}{\input{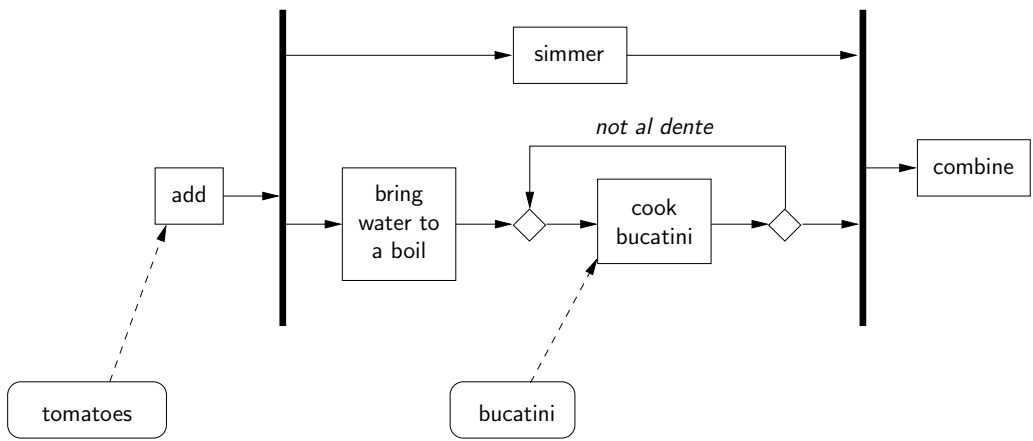}}
  }
   	\caption{An excerpt from the recipe of figure~\ref{fig:txt-recipe} as a workflow. In addition to sequences, a conjunction expressing simultaneous actions and a loop expressing an action continued until a condition is met are shown.}
   	\label{fig:form-eg-wf}
	\end{center}
\end{figure}

\begin{figure}
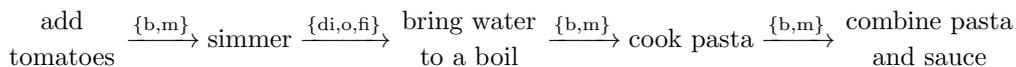

	\begin{center}
		\[
			\genfrac{}{}{0pt}{}{\text{add}}{\text{tomatoes}}
			\xrightarrow{\alrel{b,m}} \text{simmer} \xrightarrow{\alrel{di,o,fi}}
			\genfrac{}{}{0pt}{}{\text{bring water}}{\text{to a boil}}
			\xrightarrow{\alrel{b,m}} \text{cook pasta} \xrightarrow{\alrel{b,m}}
			\genfrac{}{}{0pt}{}{\text{combine pasta}}{\text{and sauce}}
		\]
   	\caption{An excerpt from the recipe of figure~\ref{fig:txt-recipe} as a set of qualitative constraints using Allen algebra (\alrelnb{di} and \alrelnb{fi} are the inverse of  \alrelnb{d} and \alrelnb{f}). The simultaneity is expressed by the relation disjunction \alrel{di,o,fi}.}
   	\label{fig:form-eg-qa}
	\end{center}
\end{figure}

Figure~\ref{fig:txt-recipe} illustrates the recipe used as a running example. An excerpt from it is represented as a workflow in figure~\ref{fig:form-eg-wf}, and as a graph representing Allen algebra constraints in figure~\ref{fig:form-eg-qa}.
One can see that workflows are able to represent the end condition on the ``cook'' action using a loop, whereas it cannot be represented in Allen algebra.\footnote{
In this instance, technically, it \emph{could} be represented if we accepted that states can be reified as intervals and thus that one can write $\text{cook pasta}\alrel{m}\text{pasta al dente}$. This is not possible for all workflow loops, however.
} On the other hand, Allen algebra represents adequately the meaning of ``meanwhile'' with the relation \alrelnb{di,o,fi}, whereas the workflow is not precise---simmering then cooking, cooking then simmering, and doing both simultaneously are all correct interpretations of this workflow.

We come back to the scientific protocol in subsection~\ref{sec:chem} to show that the same algorithms can deal with it satisfactorily.

\section{Case-based reasoning and adaptation}
\label{sec:appl}

This section briefly introduces CBR and presents two reasoning tasks in which we use the acquired cases.
The purpose is to detail what types of information must appear in the case representation in order to make those tasks possible,
and thereby show that highly structured case representations are indeed required, and cannot be acquired using information retrieval or, it seems to us, information extraction techniques.

Both tasks fit within the framework of \taaable~\cite{taaable08,taaable09,taaable10,taaable11},
  a CBR system conceived to resolve cooking problems,
  and a contestant in the \emph{Computer Cooking Contest}\footnote{\url{http://computercookingcontest.net/}}
  held annually at the \emph{International Conference on Case-Based Reasoning} since 2008.
User queries such as ``I want a dessert with rice and figs'' are addressed by retrieving one or more recipes that fit the criteria from a recipe book.
If none exists, a similar recipe is found and an ingredient substitution is suggested: ``Cook glutinous rice with mangoes but substitute figs for mangoes.''

CBR systems implement four steps, known as ``the four Rs'': retrieve, reuse, revise, and retain~\cite{lopez06ker}.
\taaable performs the \emph{retrieve} task when it finds a similar recipe,
  and the \emph{reuse} task by suggesting that the answer to the user's query is the retrieved recipe with an ingredient substitution.
It is possible, though, to go further in terms of reuse, and to actually modify the retrieved recipe so that it works with the new ingredients.
In this work, we are mostly concerned with adaptation at two interleaved aspects:
  adaptation of the preparation instructions in their formal form,
  and adaptation of the instruction text itself.
The two reasoning tasks we present are therefore adaptation tasks: one on a workflow-like graph formalism, and one on interval-algebraic constraints.

\subsection{Adaptation using grafting}
\label{sec:graft}

In~\cite{dufour10iccbr}, we proposed a method to adapt cases represented in a workflow-like formalism centred around the data stream (ingredients in this instance) rather than the actions. To adapt a recipe in which the CBR inference engine had suggested replacing an ingredient $\alpha$ with a new ingredient $\beta$, the processing specific to $\alpha$ is analysed with respect to all recipes containing $\beta$ in order to find the processing of $\beta$ closest to the original.

The idea behind this work is that there exist certain ``typical uses'' of any given ingredients, which can be extracted from the recipe book and used as adaptation knowledge. A typical use is simply a sequence of cooking actions applied to an ingredient in one recipe. When an ingredient $\alpha$ is replaced with an ingredient $\beta$, the sequence of actions used to prepare $\alpha$ must be replaced with some typical use of $\beta$.

For instance, the mango from the glutinous rice recipe is peeled, sliced and pitted. This sequence of actions is clearly not appropriate for figs. Thus, the action sequences for fig in all available recipes are extracted, and compared to the mango sequence. The most similar sequence is retained because we expect that the textual adaptation is less risky if the adaptation effort is reduced.
A piece of the retrieved case is pruned, and a piece from an adaptation case is grafted in its place, hence the name.
The same pruning--grafting process is used both at the formal and at the text levels.

For this method to work, the exact sequence of cooking actions applied to each ingredient must be known. Using information retrieval would not be satisfactory: while actions and ingredients would be known, one couldn't know which actions were applied to each ingredients. Information extraction does not seem practical either, since ingredients are not mentioned by name close to each verb that applies to them---indeed, the name of each ingredient appears at best once in most recipe preparation instructions.

\subsection{Adaptation based on belief revision}
\label{sec:br}

Cojan and Lieber propose an approach to adaptation based on revision theory. Let $\psi$ and $\mu$ be two knowledge bases~\cite{lieber07iccbr,cojan08iccbr}. The revision of $\psi$ by $\mu$ intuitively consists in (1) taking their conjunction and (2) repairing it to restore consistency, by minimally modifying pieces of knowledge from $\psi$. Revision-based adaptation consists in revising the case to be adapted (with respect to domain knowledge) by the user query (also with respect to domain knowledge). Thus, it can be seen as an ``adaptation by repair'' process.

The principle of revision-based adaptation has been applied to cases, including recipes, represented using a qualitative algebra~\cite{dufour12iccbr}.
The ingredient substitution is first blindly applied to the preparation of the recipe to be adapted,
  with great chances of leading to a formalised preparation that is inconsistent with the domain knowledge (e.g. the preparation asks to pit the figs).
Then, a repair process (based on a revision operator) is triggered in order to obtain a consistent preparation satisfying the query.
As future work, the authors intend to examine the feasibility of applying this adaptation method to workflows and to a combined workflow--interval algebra formalism.

Again, for this method to work, information about the processing of ingredients that seems impractical to obtain through information extraction is required.
For instance, the heuristic used to resolve the scope of the conjunction introduced by the adverb ``meanwhile'',
  which will be presented in section~\ref{sec:acquis},
  requires the system to know when ingredients are mixed together as the effect of a cooking action.

\section{Extraction process}
\label{sec:extr}

This section goes into the details of the case acquisition process.

The first steps are the same as in any natural language understanding system, going from the segmentation of the text into words to the identification of verb complements.
Subsection~\ref{sec:stand} quickly goes through those steps. It references some commonly used tools, and especially discusses the ways in which they must be adapted to deal with issues that are specific to process description texts.

Taking the output of those first steps as input, subsection~\ref{sec:anaph} proposes a way to resolve anaphoras, i.e. associate words from the text with the objects they are referring to, which is mostly specific to assembly instruction texts.
Then, using the actions, objects, and temporal clues identified by the previous steps, subsection~\ref{sec:acquis} details the process through which the structured case representation itself is built.
Finally, subsection~\ref{sec:adapt} discusses some special concerns that must be taken into consideration at the extraction step in order to limit the computational complexity of the adaptation step of the CBR system.

The process is illustrated on recipe texts throughout. In order to demonstrate that the same approach can be applied to different types of procedural texts, subsection~\ref{sec:chem} briefly illustrates the whole process on the scientific experimental protocol shown in figure~\ref{fig:txt-biochem}.

\subsection{Standard NLP toolchain}
\label{sec:stand}

This subsection goes through the standard steps taken by any natural language understanding system (NLU). The standard textbook for NLU is Jurafsky \& Martin's~\cite{jm09}. Several open-source tools that can greatly reduce the human cost of developing an NLU system exist. While the Stanford CoreNLP toolsuite\footnote{\url{http://nlp.stanford.edu/software/corenlp.shtml}} offers state-of-the-art performance for general purpose text, frameworks such as GATE\footnote{\url{http://gate.ac.uk/}} or OpenNLP\footnote{\url{http://opennlp.apache.org/}} offer the best flexibility.

\subsubsection{Segmentation}
The very first step in processing unstructured text is to segment it into lexical items, which are more or less equivalent to words.
This process, named \emph{tokenization}, is simple for languages which separate words with spaces, though not as simple as it may appear: for instance, ``don't'' is really two lexical items, while ``New Jersey'' is really one.

Whereas, for a language such as Chinese, tokenization is one of the biggest NLP problems, in English, regular expressions implemented by deterministic finite-state automata are sufficient to deal with it. Ready-to-use tokenizers exist for many languages (e.g. the Stanford English Tokenizer\footnote{\url{http://nlp.stanford.edu/software/tokenizer.shtml}}) which give quite satisfactory results.

At this stage, the text is also segmented in sentences using the punctuation as a guide. Domain specific \emph{named entity recognition} can be performed: for instance, in the first sentence of the recipe in figure~\ref{fig:txt-recipe}, ``olive oil'' corresponds to one ingredient and could thus be grouped as one lexical item and annotated with respect to an ingredient ontology.

The first sentence of the example recipe, after tokenization, becomes: ``Sauté -- garlic -- in -- olive\_oil -- .''

\subsubsection{Morphological analysis}
\emph{Part-of-speech tagging} is the process through which each lexical item is tagged with the appropriate class. Depending on the language considered, different techniques are more or less appropriate. An inflectionally rich language such as Russian, in which a word can have numerous forms, is not tagged in the same way as a language such as English, in which it is for instance very usual for a related verb and noun to have the same form.

Morphologically poor languages such as English are tagged based on the likelihood of tag sequences as learnt from tagged corpora. Compare for instance the two sentences ``Cream the butter'' and ``Butter the cream'': there is no information inherent to the words ``cream'' and ``butter'' that can be used to infer whether they are being used as a verb or as a noun, but the determiner ``the'' makes it clear which is a noun, making the other the verb.

General purpose, ready-to-use part-of-speech taggers exist, the most well-known being the Stanford Tagger\footnote{
	\url{http://nlp.stanford.edu/software/tagger.shtml}
}~\cite{toutanova03acl}, but they are not always useful when working with texts describing processes, mainly because those tend to use imperative forms, which are not so frequent in general purpose training corpora. Observe for instance the fact that verbs are unlikely to begin a sentence in normal discourse, while this is generally the case in imperative sentences. Consequently, it is preferable if the tagger can be trained on a corpus of the same type of texts that will need to be tagged.

For our implementation, we used a purpose-made annotated corpus consisting of 82 recipes, making 12,125 lexical items, to train a transformation-based tagger, also known as a Brill tagger. A Brill tagger first tags each lexical item with its most likely part-of-speech independent of context, then applies increasingly more specific context-sensitive rules to adjust for context~\cite{brill93phd}. A typical rule likely to be induced in any corpus would be ``retag any verb appearing after a determiner as a noun''. The first rule induced from our corpus is ``retag any noun appearing at the beginning of a sentence as a verb''.

Our tagger attains 93.1\% accuracy, significantly under the 97.5\% state-of-the-art claimed by Søgaard~\cite{sogaard11hlt}, but 32\% better than the same tagger trained with a general-purpose corpus.

The first sentence of the example recipe is tagged by our system as: ``Sauté/verb garlic/noun in/preposition olive\_oil/noun ./punctuation''.\footnote{This is a simplification obtained by grouping together different parts-of-speech. The Penn Treebank tagset, for instance, recognises 45 parts-of-speech, including 4 for nouns and 6 for verbs.} The Stanford Tagger trained on a general-purpose corpus mistakenly tags ``Sauté'' as a noun.

\subsubsection{Syntactic analysis}

The next step is \emph{parsing.} There are many ways to parse natural language text, depending upon the required results. Among the most effective parsers in existence today are the probabilistic context-free parsers (which give one and only one parse for any sentence) and the link parser (which gives as many parses as there are possible interpretations of the syntax of a sentence, and possibly none). A natural language parser, as a compiler, iteratively groups lexical items within phrases, until a whole sentence is grouped as one phrase, yielding a parse tree.

Parsing in \taaable is required only to identify the complements of verbs (which are likely to be the arguments of the actions) and the modifiers (which are likely to identify stopping conditions on the actions). Considering this, a complete parsing is not required. Indeed, since the text can be parsed clause by clause and we do not need access to such information as knowing, for instance, which noun phrase a prepositional phrase is related to, our grammar does not need recursivity. 

This means that a simpler process named \emph{chunking,} which is equivalent in complexity to a finite-state automaton, is sufficient, and a very simple grammar can be provided. For instance, here is the grammar for identifying noun phrases, formulated as pseudo-regular expressions:
\begin{itemize}
	\setlength{\itemsep}{0pt}
	\setlength{\parskip}{0pt}
	\item noun = singular\_noun \regor plural\_noun
	\item noun\_modifier = adjective \regor past\_participle
	\item determiner = article \regor numeral \regor WH-determiner\footnote{``Which'', ``whose'', ``that'', etc.}
	\item noun\_subphrase = pre-determiner$^?$\footnote{``All'', ``both'', etc.} determiner$^?$ noun\_modifier$^*$ noun$^+$
	\item noun\_phrase = noun\_subphrase ( (comma noun\_subphrase)$^*$ (comma$^?$ conjunction) noun\_subphrase)$^?$
\end{itemize}

A full parser would build the following syntax tree for the first sentence in the example recipe:
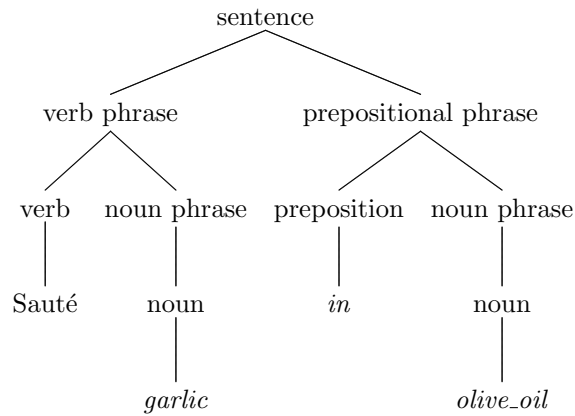
\begin{center}
	\synttree[ sentence [ verb phrase [ verb [ \text{Sauté} ] ] [ noun phrase [ noun [ \textit{garlic} ]]] ] [ prepositional phrase [preposition[\textit{in}]] [ noun phrase [ noun [ \textit{olive\_oil} ]]] ] ] 
\end{center}

Our chunker, however, will give a simplified structure which still contains all the information required by the algorithm:
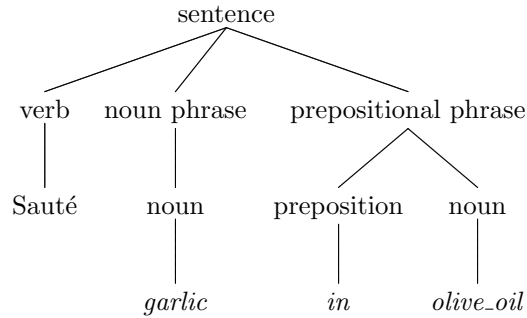
\begin{center}
	\synttree[ sentence [ verb [ \text{Sauté} ] ] [ noun phrase [ noun [ \textit{garlic} ]] ] [ prepositional phrase [preposition[\textit{in}]] [ noun [ \textit{olive\_oil} ]]] ]
\end{center}

The depth of the ``chunk tree'' is strictly limited to 3 (root, phrases, part-of-speech tags and lexical items), whereas the depth of a regular parse tree grows with the number of lexical items.

Because procedural texts normally use simple sentence structures, the output of the chunker, the punctuation and the conjunctions are sufficient to allow for clause splitting, further dividing the text so that each finite verb\footnote{
	Non-finite verbs are typically those in the participle or indicative mood, e.g. ``Fry 10 minutes, \emph{stirring} continually.'' Finite verbs include verbs in the indicative and the imperative moods.
}
sits in its own group. For instance the sentence ``Stir in pancetta and cook 5 minutes'' is divided in two clauses by the conjunction ``and'': one centred around the verb ``stir'' and the other around the verb ``cook''. In language that use them, copular verbs (e.g. ``to be'', ``to become'') must be discarded in order to obtain a single clause, i.e. ``until the garlic becomes brown'' would be parsed as ``until the garlic brown''.

\subsection{Anaphora resolution}
\label{sec:anaph}

At this stage, each verb's complements and modifiers can be identified:
\begin{itemize}
	\setlength{\itemsep}{0pt}
	\setlength{\parskip}{0pt}
	\item Any noun phrase located after the verb and referring to available ingredients is considered as an object complement of the verb.
	\item Any prepositional phrase referring to available ingredients is considered as a prepositional complement of the verb.
	\item Any prepositional phrase referring to ingredient properties (e.g. ``until brown'') is considered as a relevant modifier of the verb.
	\item Any prepositional or noun phrase referring to a duration (e.g. ``5 minutes'' or ``for 5 minutes'') is also considered as a relevant modifier of the verb.
\end{itemize}

Complements are thus known at a syntactic (surface) level, but this is not always sufficient to know which actual ingredients they are referring to.
Mapping words to actual objects presents some special difficulties for instruction texts, which make heavy use of different types of \emph{anaphoras}:
\begin{description}
	\item[Lexical anaphoras]
			A noun is used to refer to some object.
			Instructions frequently exhibit a little-studied phenomenon known as \emph{evolutive anaphora,}
			in which a word is used to refer to an object that exists at a given time, but not at another one,
			e.g. ``Add tomatoes and simmer for 30 minutes [\dots] Combine pasta and \emph{sauce}'', where ``sauce'' references the result of the ``simmer'' action.
	\item[Grammatical anaphoras]
			In common texts, those are often pronouns.
			In instructions though, tedious repetitions are often avoided by removing a verb's complement altogether,
			yielding a \emph{zero-anaphora},
			e.g. ``Sauté garlic in olive oil. Cook until brown [implicitly: \emph{the garlic}].''
			Zero-anaphoras are called that way because it is postulated that they exist in the sentence as pronouns having no realisation.
			Thus, the syntax tree of ``Cook until brown'' could be:
            \begin{center}
            	\synttree[ sentence [ verb phrase  [ verb [\textit{Cook}] ] [ noun phrase [ $\emptyset$ ] ] ] [ prepositional phrase [ preposition [ \textit{until} ] ] [ adjective [ \textit{brown} ] ] ] ]
            \end{center}
\end{description}

Typically, NLP systems solve anaphoras in three steps: first finding candidates referents, second filtering the list with grammatical criteria, and finally selecting one using varied heuristics. The first step can be made much easier by keeping an up-to-date set of available objects that can be passed as arguments to actions. The second step is not relevant for the present application because the anaphoras most often contain no grammatical clues about their referent. The third step is solved using a very simple heuristic. In cooking, the objects referred to by anaphoras are  components: raw ingredients that are part of the set of foods created by previous actions.

\subsubsection{Lexical anaphoras}
To solve evolutive anaphoras, therefore we keep a set of available food components, called \emph{domain,} up-to-date. It is initialised with the ingredients from the ingredients listings. The food components are considered as consumable resources, i.e. an action will remove some food components from the domain and add some others. For instance, the clause ``mix flour, eggs and milk'' would remove the three food components associated to flour, eggs and milk from the domain, and would add a new food component in it, which becomes available to the clause ``pour batter''. Having an up-to-date domain makes it simpler to identify the object(s) an evolutive anaphora is referring to.

The domain of food components available at step $s$ is called $\mathcal{D}_s$: $\mathcal{D}_0$ is the initial domain, and $\mathcal{D}_{s+1}$ is the domain after the $s$th action in the sequential order of the text has been analysed. The domain is ordered with respect to the order of the ingredients list, which is necessary to resolve anaphoras such as ``the next 5 ingredients''. The order of food components added subsequently is thus not important. The objects in the domain are called ``food components''. A function $\mathcal{I}$ is defined which, for a given food component $F$, gives the set $\mathcal{I}(F)$ of its ingredients.

Depending on how the domain is searched for the correct referent of an anaphora, two types are distinguished: existential and universal references.

\paragraph{Existential references} An existential reference is so called because it captures a food component which contains \emph{some} specific ingredient. The ingredient is sometimes mentioned explicitly, making the reference trivial to solve, e.g. a ``beef mixture'' can be whatever food component contains beef. Anaphoras such as ``batter'' are harder to solve. 

The NLP process described in the previous subsection was used to identify all the nouns in a corpus of recipes that could not be associated to a listed ingredient, most of which were indeed lexical anaphoras. Then, the set of ingredients referred to in each instance was manually built. This makes it possible to show, for instance, that a ``batter'' contains at least eggs or flour in over 99\% of recipes from a given recipe book. This yields a target set of ingredients expected in a food component, called $\mathcal{T}$.\footnote{
	On a very large corpus, it may be possible to automate this process by setting a threshold and considering that \emph{all} ingredients that have been mixed with others at the point where the anaphora is met are candidates for the target set.
}

We consider that any food component in the domain that contains at least one ingredient in the target set is the food component referred to. In practice, there is almost always only one, but heuristics could be designed to choose one in other cases, such as taking the one with the \emph{most} ingredients from the target set.

The set of food components that could be the referent to an existential anaphora with target set $\mathcal{T}$ at step $s$ is defined by:
\begin{equation}
       \{ F \in \mathcal{D}_s \mathrel{|} \exists i , i \in \ing(F) \land i \in \mathcal{T} \}
\end{equation}

\paragraph{Universal references} A universal reference is so called because it captures a set of foods for which \emph{all} ingredients are of a certain type. This is the case, for instance, when a recipe says to ``sift all dry ingredients together''. Given that an ontology exists and that it is possible to map the anaphoric reference to a concept of this ontology, solving a universal reference is straightforward. For instance, ``dry ingredients'' represents the set of foods which are exclusively made of ingredients appearing under the ``dry ingredient'' concept in the ontology.

The set of food components that a universal anaphora pointing to ontology class $\mathtt{C}$ refers to at step $s$ is defined by\footnote{
	Where $i$ is both an ingredient in the recipe and a concept in the ontology corresponding to this ingredient, an abuse of notation which increases readability. $i \sqsubseteq \mathtt{C}$ means that $i$ is a subclass of $\mathtt{C}$.
}:
\begin{equation}
		 \{ F \in \mathcal{D}_s \mathrel{|} \forall i , i \in \ing(F) \to i \sqsubseteq \mathtt{C} \}
\end{equation}

\subsubsection{Grammatical anaphoras}
The main kind of grammatical anaphora used in instruction texts is zero-anaphoras. Their resolution, while a complicated problem in general, can usually be dealt in instructions with simple heuristics. As discussed earlier, most systems need to build a list of candidate referents, though this is not necessary in this instance with the use of the domain, then filter it, which is impossible in the case of zero-anaphoras. The last step is to select one referent using preference heuristics.

The one heuristic all theories seem to agree on is recency: all other things being equal, the most recently introduced entity among the candidate referents is selected. This simple heuristic is usually sufficient in instruction texts, with a small adjustment to fit within our evolutive domain model: a zero-anaphora is a reference to the objects that were inserted in the domain by the last action that occurred before the current clause.

The most important problem they pose is their detection. A human would instinctively know that something is missing from the instruction ``add milk'' (i.e. ``to what?''). For the computer to detect this implies a high level of linguistic knowledge. The fact that an argument is missing could be found out using either syntactic or semantic frames, the former being a much easier option to implement while the latter is less error-prone. Subcategorization frames specify, for a given verb, the types of complements that must accompany it, e.g. ``add needs an object and a prepositional complement: add $O$ to $P$'', making it obvious when one is missing.
Subcategorization frames can be acquired automatically in corpora, or taken from online databanks such as VerbNet~\cite{schuler05phd}, but in any case it is advisable to verify manually that they are fit for a given application (there may be surprisingly few anyway, e.g. only 130 in our cooking application).

If the anaphora was an actual pronoun instead of being a zero-anaphora, e.g. if the second sentence of the example recipe read ``Cook \emph{it} until brown'', the result would be the same. Therefore, it is easiest to disregard the pronouns altogether, which will cause the system to think a complement is missing and use the heuristic for zero-anaphoras.

\subsubsection{Updating the domain}
After resolving the anaphoras, the \emph{actual} arguments of the action at step $s$ are known.
This subsection shows how the domain of food components available at step $s+1$ is computed from the arguments, with respect to the nature of the action.

Most actions take sets of food components as arguments. Those will be identified hereafter with capital letters, whereas sets of simple ingredients will be identified with lower case letters. The set of food components which appear as an object complement of the verb is called $O$, and the set of food components which appear as a prepositional complement is called $P$. Certain verbs may take a set of \emph{ingredients} as an object rather than a set of food components. This is the case, for instance, of remove: ``remove garlic [implicitly: from the mixture in which it is]''. In this case, the set of ingredients is identifies with a lower case $o$.

While we consider that any food component passed as an argument to any verb is consumed by the associated action, deciding which food components are output by the action is less straightforward. Four different classes of actions are distinguished according to the food components they output. In the following, each output food component is represented as $N$. When there is more than one (this is defined by the action class), the set of $N$s is called $S$.
\begin{description}
	\item[Union actions]
		Output only one object, the parts of which are the union of the parts of all the inputs, e.g. ``Combine pasta and sauce''.
   		\begin{equation}
         	\begin{split}
                \mathcal{D}_s =& ( \mathcal{D}_{s-1} \backslash ( O \cup P ) ) \cup \{N\}, \quad \text{with}\\
                & \ing(N) = \ing(O) \cup \ing(P)
            \end{split}
         \end{equation}
	\item[Parallel actions]
		Output as many objects as were input, implying that the action, while expressed as one clause, is executed separately for each input, e.g. ``Mince the onions, the carrots and the celery''.\footnote{
			In equation~\ref{eq:parallel}, $\exists!$ represents uniqueness quantification, i.e. $\exists!x,P(x)$ is defined as
			$\exists x,P(x) \land \forall y, P(y) \to x=y$.
			}
         \begin{equation}\label{eq:parallel}
         	\begin{split}
                \mathcal{D}_s =& ( \mathcal{D}_{s-1} \backslash O ) \cup S,  \quad \text{where}\\
                & \forall F , F \in O \to \exists! N , N \in S \land \ing(N) = \ing(F)
            \end{split}
         \end{equation}
	\item[Splitting actions]
		Input only one object and output several, e.g. ``separate the egg's yolk from the white''.
         \begin{equation}
         	\begin{split}
                \mathcal{D}_s =& ( \mathcal{D}_{s-1} \backslash O ) \cup S,  \quad \text{with}\\
                & \ing(O) = \bigcup_{N \in S} \ing(N) 
            \end{split}
         \end{equation}
	\item[Difference actions]
		Output only one object, the parts of which are the parts of its prepositional complement input minus the part specified as its object, e.g. ``remove the garlic from the oil''. Difference actions are distinguished from splitting action from a practical point of view: the removed part is discarded from the domain.
         \begin{equation}
         	\begin{split}
                \mathcal{D}_s =& ( \mathcal{D}_{s-1} \backslash P ) \cup \{N\},  \quad \text{with}\\
                & \ing(N) = \ing(P) \backslash o
            \end{split}
         \end{equation}
\end{description}

\subsection{Case acquisition}
\label{sec:acquis}

Because instruction texts are usually designed such that the instructions are to be applied in the textual order of their description, the task of extracting a workflow from them is not very complicated: once all the actions described in the text have been fully analysed and the verb modifiers are known, generating a workflow primarily consists in applying a mapping from the set of modifiers to the set of workflow structure patterns.
This last step can be implemented so as to operate iteratively at the same time as the anaphora resolution process. Because it is partly dependent on the ingredients, though, it is easier to describe assuming that all anaphoras are resolved.

\begin{figure}
	\begin{center}
  {
  \def\mef#1{{\huge$#1$}} 
  \def\meft#1{\text{#1}} 
  \def\saute{\meft{\mlm{saut\'e garlic\\ in olive oil}}}
  \def\cook{\meft{\mlm{cook until\\ brown}}}
  \def\discard{\meft{\mlm{discard\\ garlic}}}
  \resizebox{0.5\textwidth}{!}{\input{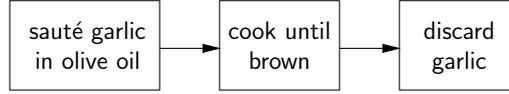}}
  }
		\caption{Extracted sequence.}
		\label{fig:acq-seq}
	\end{center}
\end{figure}

Two actions textually juxtaposed without further precision are considered as taking place sequentially. Therefore, disregarding the ``until brown'' condition for now, the first two sentences of the example recipe of figure~\ref{fig:txt-recipe} are represented as shown in figure~\ref{fig:acq-seq}. In interval algebra, the disjunctive relation \alrel{b,m} can express a sequence.

\begin{figure}
	\begin{center}
  {
  \def\mef#1{{\huge$#1$}} 
  \def\meft#1{\text{#1}} 
  \def\immerse{\meft{immerse peppers in hot water}}
  \def\donothing{\meft{do nothing}}
  \def\drypeppers{\meft{\emph{using dry peppers}}}
  \def\freshpeppers{\meft{\emph{using fresh peppers}}}
  \resizebox{0.75\textwidth}{!}{\input{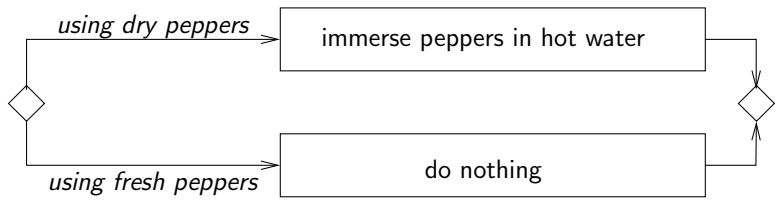}}
  }
		\caption{Extracted disjunction.}
		\label{fig:acq-disj}
	\end{center}
\end{figure}

Explicit disjunctions in instruction texts are rare. They usually stem from a decision that is made before the procedure begins. In recipes, for instance, most disjunctions come from ingredient options, e.g. using either dried or fresh peppers. In this context, it is usually enough to say that an action is in a disjunction if it applies to an ingredient that is itself in a disjunction. For instance, sentence ``If using dry peppers, immerse them in hot water'' can be represented as in figure~\ref{fig:acq-disj} without having fully to analyse the phrase ``If using dry peppers''---the condition can be inferred from the ingredient disjunction.

\begin{figure}
	\begin{center}
  {
  \def\mef#1{{\huge$#1$}} 
  \def\meft#1{\text{#1}} 
  \def\cookPasta{\meft{cook pasta}}
  \def\doNothing{\meft{do nothing}}
  \def\combine{\meft{\mlm{combine pasta\\ and sauce}}}
  \def\pastaNotAlDente{\meft{\emph{pasta not al dente}}}
  \def\pastaAlDente{\meft{\emph{pasta al dente}}}
  \resizebox{.9\textwidth}{!}{\input{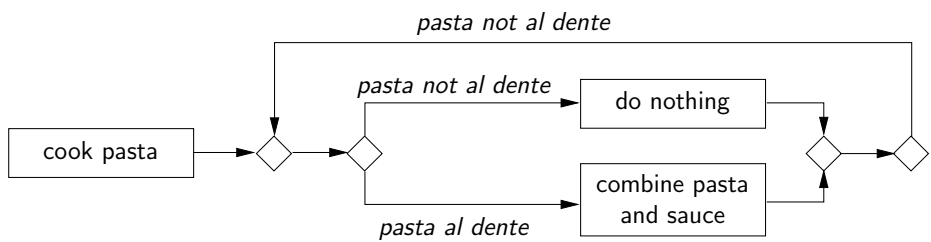}}
  }
		\caption{Extracted loop with short scope.}
		\label{fig:acq-loop}
	\end{center}
\end{figure}

More frequently, disjunctions are used inside loops to specify a condition under which an action can be stopped, as in ``Cook pasta until al dente.'' We have chosen to represent this as a single task with an explicitly specified end condition, but it may easily be represented as in figure~\ref{fig:acq-loop} since the condition ``until al dente'' is identified at the chunking step and it can be inferred that the property applies to the input of the action. The same can be represented in interval algebra by considering that ``al dente'' is a state of ``pasta'' which is reified as an interval, such that the ``cook pasta'' interval is in \alrelnb{m} relation with the ``pasta al dente'' interval.

Loops are also used occasionally for true repetitions, triggered by the verb ``repeat''. While the end conditions will there again be specified as a verb modifier, the difficulty is determining the scope of the repetition. We propose a simple heuristic: if the verb ``repeat'' has a prepositional complement $P$ and there previously was a splitting action at step $s$, such that $P$ can be resolved to a food component in $D_s$, then the loop is scoped over the whole sequence from the splitting action to just before the ``repeat''. In all other cases, either there are cooking actions in the same sentence as ``repeat'' and the loop is scoped over those, or then the loop is scoped over the actions of the previous sentence. An example of the latter case would be: ``Add a spoonful of water and knead. Repeat until all flour is absorbed.'' Loops can also be used with sporadic repetitions, such as in ``stir frequently''. In this case, the action is inserted in a sequence with a ``do nothing'' action, and this sequence is looped. This normally appears as an action simultaneous with another.

\begin{figure}
	\begin{center}
  {
  \def\mef#1{{\huge$#1$}} 
  \def\meft#1{\text{#1}} 
  \def\fryVegetables{\meft{fry vegetables}}
  \def\stirVegetables{\meft{stir vegetables}}
  \resizebox{0.4\textwidth}{!}{\input{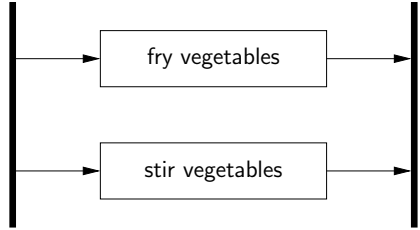}}
  }
		\caption{Extracted conjunction with short scope.}
		\label{fig:acq-conj1}
	\end{center}
\end{figure}

The conjunction workflow structure is used to represent a possible simultaneity. This can be triggered by verb modifiers such as ``meanwhile'', but also simply by the use of the participle mood, e.g. ``Fry 10 minutes, stirring constantly.'' The latter case yields a conjunction that is scoped only over the previous action, as shown in figure~\ref{fig:acq-conj1}. In interval algebra, while this does not have the exact same semantics, it would make sense to represent this kind of simultaneity as inclusion using the relation \alrelnb{d}.

\begin{figure}
	\begin{center}
  {
  \def\mef#1{{\huge$#1$}} 
  \def\meft#1{\text{#1}} 
  \def\cook{\meft{\mlm{cook pasta\\ until \\al dente}}}
  \def\simmer{\meft{simmer}}
  \def\salt{\meft{salt}}
  \def\add{\meft{\mlm{add\\ tomatoes}}}
  \def\combine{\meft{\mlm{combine\\ pasta and\\ sauce}}}
  \def\boil{\meft{\mlm{bring\\ water to\\ a boil}}}
  \resizebox{\textwidth}{!}{\input{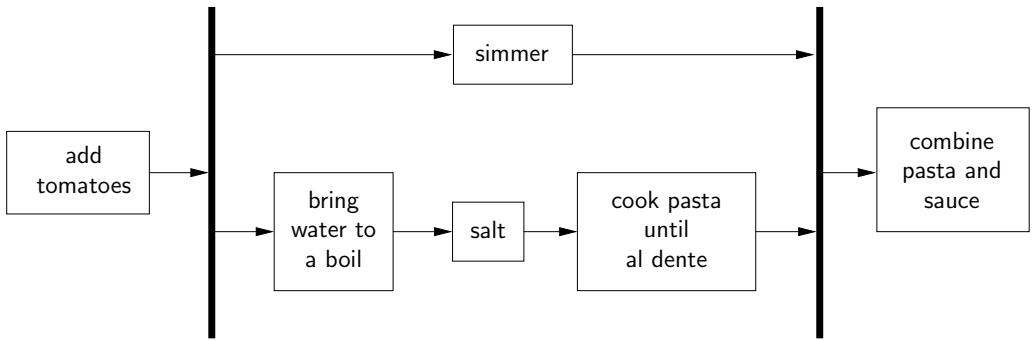}}
  }
		\caption{Extracted conjunction with long scope.}
		\label{fig:acq-conj2}
	\end{center}
\end{figure}

``Meanwhile''-like modifiers are more tricky: they initiate a sequence of actions that can be simultaneous with the preceding action. In this case, we consider that the conjunction extends all the way to just before a step on which any food component output by an action in the sequence serves as input to the same union action as any food component output from the previous action. For instance, the action immediately previous to ``Meanwhile, bring water to a boil'' in the example recipe is ``simmer'', which has the sauce for output. Therefore, the conjunction extends to just before ``Combine pasta and sauce'', as shown in figure~\ref{fig:acq-conj2}. In interval algebra, this is more straightforward, as the relation \alrel{d,oi,f}$^?$ between the first action of the simultaneous sequence and the previous action is sufficient---the other relations can be inferred.

Some temporal phenomena that can be represented in interval algebra but not with workflows may also be of interest for reasoning.
It is expected that those are also to be found in verb modifiers, and should be pretty easy to analyse.
In the example recipes, there are two different types: single action durations such as ``Fry 10 minutes'', and relations between actions such as ``Serve immediately.''
The latter can be represented simply using the \alrelnb{m}$^?$ relation, whereas the former requires the time to be reified by creating an interval whose gist is to last 10 minutes and the use of the \alrelnb{$?^=$} relation from \indu.

Workflows can also represent the dataflow. In the cooking domain, the dataflow is the ingredients input and output by each task. This information is computed as part of the anaphora resolution process of subsection~\ref{sec:anaph}, so it is straightforward to integrate it within the generated workflow.

\subsection{Annotating for text adaptation}
\label{sec:adapt}

The methodology presented thus far makes the automatic acquisition of cases representing processes possible using a formalism similar to those most often used in the domain. But this is still insufficient to allow for solid textual adaptation---that is, for instance, to allow \taaable to answer a user query with an adapted recipe text generated on the fly.
This subsection presents ongoing work about the problem of annotating the text during case acquisition in order to make the textual adaptation process easier.

Whatever the case representation formalism, the effect of any adaptation method will be to modify the formal representation of the retrieved case to make it suitable as a solution to the target problem. If the solution is to be presented to the user in text format, it is necessary that these modifications be applied to the text at the same time as to its formalised form.

In section~\ref{sec:graft}, the adaptation method that consists in replacing sentences applicable to an ingredient $\alpha$ with sentences more suited to a substitution ingredient $\beta$ developed in~\cite{dufour10iccbr} was presented briefly.
Compared with, for instance, complex and error-prone text generation, this method is very economical in terms of textual adaptation, because it makes it possible to prune a piece of text and replace it with a piece from a different text, modulo minor linguistic adjustments (fixing capitalisation and punctuation), through a parallel process.

It still requires an annotation of the text parallel to the extraction of its formal representation. Minimally, the clause segmentation must be marked in the text, and each clause must be annotated with the ingredients that its action take as input. In order simply to replace whatever $\alpha$-specific preparation steps in the recipe with $\beta$-specific preparation steps, this is sufficient.

A more thorough type of adaptation, such as the one proposed by Minor et al.~\cite{minor10iccbr}, where a workflow representation is edited following a set of precise instructions, would require a more thorough annotation of the text. It would require not only the workflow tasks to be cross-referenced with the clauses, but also the control flow structures to be cross-referenced with the phrases that materialise them in text, requiring a full temporal annotation language.

TimeML~\cite{pustejovsky04lot} seems to be the most expressive language of this kind in existence. It makes it possible to represent, directly in text, the relations between actions, states, intervals, time points, and even sets of time points. Figure~\ref{fig:timeml} shows one possible TimeML of the instructions ``Add tomatoes and simmer 30 minutes. Meanwhile, bring water to a boil'' in TimeML. Tasks in a process are represented as \texttt{EVENT}s. \texttt{TIMEX3} tags encode explicit temporal expressions that are related to an \texttt{EVENT} and specify the time of their occurrence or their duration, whereas \texttt{SIGNAL} tags underline words or phrases that are used to indicate the temporal relations between events.

\begin{figure}
	\input{timeml}
	\caption{An excerpt from the recipe of figure~\ref{fig:txt-recipe} annotated with TimeML.}
	\label{fig:timeml}
\end{figure}

Those relations are annotated using the \texttt{TLINK} tag with a \texttt{relType} attribute that offers values comparable to the relations of interval algebras.
While TimeML can be used as a temporal knowledge representation formalism in its own right, it can also be used merely as a format to materially represent temporal annotations in a different formalism in text, which is what the authors propose.
Figure~\ref{fig:timeml} shows a ``regular'' TimeML representation of the recipe excerpt of figure~\ref{fig:txt-recipe}, but it would be straightforward to replace the \texttt{relType} values with \indu relations.
It would be perfectly feasible as well, given a workflow expressed as XML, to cross-reference an \texttt{EVENT} with a workflow task, as well as a \texttt{SIGNAL} with a workflow with a workflow control structure. The text parts impacted by any modification of the workflow would thereby be identified easily.

For workflow or algebraic adaptation as presented is section~\ref{sec:br}, though, linguistic knowledge complementary to that described in the previous subsection is required. Indeed, in section~\ref{sec:acquis}, we described the case acquisition as little more than applying a mapping from verb modifiers (\texttt{TIMEX}s and \texttt{SIGNAL}s) to workflow structure patterns (or interval algebra relations). Text adaptation that is not purely grafting-based imply a measure of text generation, which requires an inverse mapping from workflow patterns or algebra relations back to verb modifiers.

As an example, if the conjunction between the ``simmer'' and the ``bring to boil'' actions, which was inferred because of the \texttt{SIGNAL} ``meanwhile'', was to be replaced by a sequence at the adaptation stage, it would become necessary to replace the word ``meanwhile'' as well to ensure the equivalence of the text and its formal representation. Since the sequence was defined as the default relation when no modifier is frequent, in this case, the ``meanwhile'' could be removed altogether from the text, modulo capitalisation and punctuation fixing.

\subsection{Application to the scientific example}
\label{sec:chem}

This section gives an overview of the additional work that would be required to make the method described above work with a different domain, that of the scientific experimental protocol shown in figure~\ref{fig:txt-biochem} on page~\pageref{fig:txt-biochem}.

As for the standard NLP toolchain, it can be applied in exactly the same way to this experimental protocol as to recipes. This very example shows some sentences that are not in the imperative, but since this also occurs in some recipes, the part-of-speech tagger trained on recipes gave the expected results on the protocol.

For the anaphora resolution though, additional resources are required: the ontology of objects, the dictionary of actions, and the target sets of lexical anaphoras. 

The ontology must contain the chemicals that may be used in the protocol, such as yeast, trypsin or Na$_2$HPO$_4$, and is used to find those objects in the text and to resolve lexical anaphoras.
The dictionary of actions must contain the relevant verbs, such as ``suspend'', ``boil'' or ``centrifuge'', along with their subcategorization frame to allow identification of zero-anaphoras, and their class for domain updating. For instance, ``suspend'' is a union action requiring both an object and a prepositional complement, and ``centrifuge'' is a splitting action requiring but an object.

Regarding lexical anaphoras needing target sets to resolve, it is less clear what may be the requirements of different application domains. The semi-automatic process used for building the target sets in the cooking domain rely on the ingredients list, but it may also be used in texts without lists provided that the anaphoras don't appear in the ontology. Otherwise, those anaphoras may be difficult to deal with.
An additional piece of information about lexical anaphoras which is required for scientific protocols is the names through which the outputs of a splitting action can be referred as. In cooking, the only obvious case in which a splitting action assigns a specific name to its outputs is the separating an egg into its yolk and white. In scientific protocols, most splitting actions assign names to their outputs irrespective of the objects consider: for instance, the outputs of ``centrifuge'' in the example protocol are ``precipitate'' or ``residue'' and ``supernatant''. Therefore the application must know that, e.g. in step 3, ``precipitate'' refers to one of the two outputs of the ``centrifuge'' action, and ``supernatant'' to the other.

Some sentences are in the passive voice and have subjects instead of objects. The subjects are not considered when looking for action arguments. This has no effect in this case, as the missing argument is interpreted as a zero-anaphora and correctly resolved. Larger scale testing of the algorithm in domains other than cooking would make it possible to say whether this can be considered as a valid heuristic or whether this is a stroke of luck. In the latter case, it would be straightforward to analyse the sentence correctly, since the output of the part-of-speech tagger makes it possible to infer whether a verb is in the passive voice.

As far as the actual case acquisition step is concerned, what is suggested is mostly sufficient to deal with the protocol shown. Because the rules were built by observing the phenomena in recipe texts, some additional rules would be required for other domains. For instance, all the durations and temporal relations are correctly identified by the algorithm, except for the parts of step 2 in which the order of the text does not respect the temporal order. In the expression ``treated as in (step a)'', if one can identify the ``step a'' and it refers to a series of actions in which each takes as input the output of the previous, as is the case, it is easy to copy those actions. Identifying the step by its number would require the implementation to remember the said number at the clause segmentation stage---the current implementation internally numbers the clauses with no respect to any numbering that may be present.

\section{Case Acquisition Evaluation}
\label{sec:eval}
A prototype workflow extractor for recipes following the algorithms presented \linebreak throughout section~\ref{sec:extr} has been implemented in Perl. This makes it possible to evaluate our extraction process against a set of 15 recipes for which a workflow was created by hand by Minor et al.~\cite{minor10ccc}.\footnote{
	Those recipes are simply the first ones alphabetically in the recipe book provided by the \emph{Computer Cooking Contest} for the adaptation challenge,
	therefore we cannot guarantee representativeness.
} The prototype is quite fast, processing a recipe in an average time of 0.01 second on a 2.5~GHz i5 CPU.

Those gold standard workflows are represented as labelled graphs as introduced by Bergmann and Gil~\cite{bg11iccbr}. We evaluated the control flow (tasks and order) and the dataflow (ingredients) separately. On average, a recipe contained 15 tasks and 12 ingredients. The results are presented in table~\ref{tab:eval}.

For the control flow, the \emph{task,} \emph{and,} \emph{xor} and \emph{loop} nodes were considered. A node present in both the gold standard and the generated workflow, in the correct place and labelled with the correct action verb, is counted as correct. A node present in the gold standard workflow but not in the generated workflow is counted as a recall error (false negative), and a node present in the generated workflow but not in the gold standard is counted as a precision error (false positive). The gold standard has no condition for \emph{xor} or \emph{loop}, therefore the proposed conditions were not evaluated.

Because different workflow structures may at times represent accurately the same text, so much so that Walter et al.~\cite{walter11ccc} suggested evaluating their workflow extraction system through user experiments, we tolerated certain specific differences\footnote{
	To avoid the possible arbitrariness of having to define such equivalences and make automatic evaluation possible, a future evaluation could operate on the workflow semantics instead, as discussed by Schumaher et al.~\cite{schumaher12www}.
}:
\begin{itemize}
	\item A \emph{task} node corresponding to an action that exists in the recipe but was not included in the gold standard for methodological reasons is discarded.
	\item A \emph{task} node corresponding to an action $A$ and specifying that the action is carried out for a given length of time or until a given state is attained is considered equivalent to the use of a \emph{loop} containing a \emph{task} node corresponding to action $A$ in the gold standard.
	\item A \emph{task} node corresponding to an action represented by one verb in the recipe text but including many ingredients is considered equivalent to an \emph{and} node that has as many \emph{task} children nodes as there are ingredients all corresponding to the same action.
\end{itemize}

The recall is the ratio of the total amount of nodes correctly generated over the total amount of nodes in the gold standard, and the precision is the ratio of the total amount of nodes correctly generated over the total amount of nodes generated.

For the dataflow, only the first use of each ingredient was predictably annotated in the gold standard, so the evaluation was based on the first appearance. An ingredient input attached to the same \emph{task} node in both the gold standard and the generated workflow is correct. An ingredient indicated in the gold standard but not in the generated workflow is a recall error. An ingredient indicated in the generated workflow but not in the gold standard is a precision error.

\begin{table}
	\begin{center}
      \begin{tabular}{lcc}
      	& Recall & Precision \\ \hline
      	Control flow & 75\% & 91\% \\ \hline
      	Dataflow & 73\% & 99\% \\ \hline
      \end{tabular}
   	\caption{Workflow extraction evaluation results.}
		\label{tab:eval}
	\end{center}
\end{table}

\begin{table}
	\begin{center}
      \begin{tabular}{lcc}
      	& Recall & Precision \\ \hline
      	Control flow & 89\% & 100\% \\ \hline
      	Dataflow & 89\% & 100\% \\ \hline
      \end{tabular}
   	\caption{Results without part-of-speech error.}
		\label{tab:eval-pos}
	\end{center}
\end{table}

When looking closely at the errors, it appears that a large part of the errors occur because of wrong part-of-speech tagging of verbs as nouns or vice versa. For instance, one recipe from the evaluation set triggered both a false negative and a false positive because the verb ``drain'' was tagged as a noun, causing an action to be missed, whereas the noun ``heat'' in ``high heat'' was tagged as a verb, causing an action to be wrongly generated.

When the software was provided hand-corrected parts-of-speech, the recall increased significantly, and all false positives disappeared, as shown in table~\ref{tab:eval-pos}.

A high score in control flow extraction indicates that the actions were correctly identified and that the appropriate workflow structures (sequences, conjunctions, disjunctions and loops) were created. A high score in dataflow extraction means that the ingredients were correctly identified and that each of them was associated with the correct action. The high precision score of our implementation shows that when it makes an annotation, it is usually right. Its lower recall score means that it sometimes omits to make a required annotation. For certain applications, it is more important to have undoubted knowledge than to have complete knowledge, making those results satisfactory. For adaptation purposes, it isn't clear which if any is more important, meaning that we should strive to improve our recall score.

At 93.1\% accuracy, the part-of-speech tagger is the weak link in the proposed NLP chain. Comparing tables~\ref{tab:eval} and~\ref{tab:eval-pos} shows that the tagger is indeed responsible for most errors. The tagger would probably perform better if it was trained using a larger corpus. A less costly way to improve the results may be to change the tagger so that, when there is high uncertainty in tagging a word as a noun or as a verb, it is allowed to output more than one tagging. Then, considering that the NLP process is fast and that the tagger applies an average of 5.7 verb-related rules in a recipe, it should be possible to carry on the analysis of all possibilities and to define a heuristic to select what looks like the best output. Considering that part-of-speech tagging is required for information extraction as well, we do not think that this problem questions the choice of using NLP.

In the control flow, most of the remaining recall error (after part-of-speech correction) is due to the ontology of actions not being complete enough. Perhaps using lexical databases such as WordNet\footnote{\url{http://wordnet.princeton.edu}} could help with this problem and even be used to make the application more easily ported to different domains. In the dataflow, most of the remaining recall error is due to ingredients being used in the recipe text without being called for in the ingredient list. This is actually an intrinsic problem to the implementation, which uses the list rather than the ontology when annotating the ingredients in the text. In our opinion, the annotation should be made using the ontology.

While this evaluation is basic, the results are encouraging.
However, there is little hope to improve the system so that it reaches $100\%$
 recall and $100\%$ precision on all English procedural texts.
For this reason, we have developed a semi-automatic procedure for
 case acquisition:
 a first version of the procedural case is automatically generated,
 then an interaction with an expert makes it possible to validate it, correcting it if needed~\cite{dufour12cwc}.

Since the NLP process roughly follows the order of the text and since an error may generate further errors,
 the expert fixes the first error (in the order of the text),
 then the NLP process taking into account the correction is triggered again,
 and the correction/validation process resumes on the new version of the automatically acquired case.

To make a larger-scale evaluation possible, two things would be required:
\begin{enumerate}
	\item A larger set of hand-created or validated workflows associated to text, of which the obtention
			will be facilitated by the semi-automatic process just described, and
	\item A framework making it possible automatically to compare workflows at a semantic level, which removes the necessity
			of manually looking for equivalent workflow structures.
\end{enumerate}

\section{Related work}
\label{sec:rw}
Ever since what can arguably be described as the foundational paper in textual CBR methodology by Brüninghaus and and Ashley~\cite{brueninghaus97iccbr}, techniques beyond information retrieval (IR) were seen as too knowledge-intensive to be practical for use with CBR.
IR approaches can as a matter of fact be entirely satisfactory for CBR systems manipulating feature-based case representations. Some systems used to acquire this type of cases are indeed quite advanced, such as Wiratunga et al.'s~\cite{wiratunga04iccbr}.
In fact, some featureless textual CBR systems have been shown to give better results than the equivalent feature-based systems, e.g. Delany and Bridge's~\cite{delany07iccbr}.

The usual bag-of-words model has also been improved, for instance using graphs to represent text~\cite{cunningham04iccbr}, taking into account the position of words, which makes it possible to identify the negation of terms. Some approaches, such as the one presented by Asiimwe et al.~\cite{asiimwe07aiis}, proposed even richer case representations, requiring the need of information extraction (IE). IE has been shown to improve significantly results in CBR, as shown in Recio~García et al.~\cite{recio05iccbr}, where a significant increase in recall in observed for the retrieval task. And indeed, IE has become more efficient over the years~\cite{brueninghaus01iccbr}.
For instance, Walter et al.~\cite{walter11ccc} proposed a simple, term-based IE approach for extracting workflows from recipes, based on hand-made rules dealing with action verbs and ingredient nouns. Schumaher~\cite{schumaher12www} implemented this proposal, as well as a more advanced one, in which verbs instantiate frames containing slots that must be filled with content extracted from text. Those frames are similar to the subcategorization frames we use to detect zero-anaphoras.

We argue that there can be a significant advantage to go further and exploit different techniques taken from NLP research. For instance, Schumaher's method is likely to be easier to implement, but it also seems to have some limitations. Because it does not deal with the anaphora problem, for instance, we expect its recall to be lower than ours. Also, not fully processing the input and output of each action will make it more complicated to compute the scope of the conjunctions and loops operating over sequences of actions. As a matter of fact, to our knowledge, no IE approach has been proposed so far to deal with those workflow structures.\footnote{
	Walter and Schumaher's approaches were tested by having a panel of human experts grade the results of their two extraction processes. This means that there is unfortunately no straightforward way for us to compare our actual results to theirs. It would be an interesting perspective to agree to a common evaluation framework for workflow extraction.
}

Additionally, while it is commonly said that the more general a domain is, the more complex the linguistic models required to process text are, the converse is also true: a textual CBR system using deep NLP techniques, such as the one proposed by Gupta and Aha~\cite{gupta04flairs}, would be able to deal with texts from any domain. While we do not go as far, we think that the case acquisition system we presented in this paper could easily be used with different types of procedural cases, given a proper ontology of the action inputs and a proper verb dictionary including subcategorization frames and action types can be provided. On the other hand, a system based on IE would additionally need a new set of extraction rules.

This paper also suggests that the case acquisition process should be designed with textual case adaptation in mind.
While natural language generation is a possibility, it has been observed that the results is absolutely not comparable in quality to text produced by a human, creating a need for text reuse~\cite{gervas07tcbr}. Foundational work on this idea was accomplished by Lamontagne and Lapalme~\cite{lamontagne04eccbr}.
We believe the main originality of our work is the fact that the tie we maintain between texts and their formal representation makes it possible to use adaptation methods not specifically designed for textual CBR to adapt text.
This however means that our framework is strictly symbolic, making most of the existing work on textual reuse difficult to apply for us.

\section{Conclusion}
\label{sec:concl}
Process-oriented case-based reasoning requires a highly structured case representation, making case engineering especially complicated and costly.
This paper introduces a method for the automatic acquisition from procedural texts, and especially instruction texts, representing processes.
Our methodology was implemented in a system to generate workflows from cooking recipes, obtaining a 75\% recall and 94\% precision performance in a basic evaluation.

Although this automatic case acquisition process
 has been developed and tested on recipes, it should also be applicable to the
 procedural texts of other domains (e.g. chemistry protocols as the
 one presented in figure~\ref{fig:txt-biochem}).
This would require an ontology of the inputs of such procedures
 (e.g. ingredients for recipes) and a dictionary of the relevant verbs containing subcategorization frame and the action type of each.

The automatic acquisition could also be improved by following on some leads developed in section~\ref{sec:eval}: improving the part-of-speech tagger or implementing a tagger with multiple output for cases with high uncertainty, improving the ontology of actions, and actually using the ontology of objects when annotating the text in the implementation. Those improvements would be easier to implement and test if a better, entirely automatic evaluation framework was designed.

This paper has focused on case acquisition, but this cannot be considered outside the context of the intended reasoning tasks. One such task that is very important to us is adaptation, including textual adaptation, which comes with certain requirements that must be taken care of at acquisition time. Reflections on this topic were discussed in section~\ref{sec:adapt}, and are expected to be the object of further work within a short time frame.


\section*{Acknowledgements}
We heartily thank Mirjam Minor, Ralph Bergmann, Sebastian Görg and Kirstin Walter for making the case base they engineered in~\cite{minor10ccc} available to us for the evaluation,
as well as the reviewers for their detailed comments, 
which we feel have helped us improve the article substantially.


\end{document}